\documentclass{article} % For LaTeX2e
\usepackage{iclr2026_conference,times}

\usepackage{amsmath,amsfonts,bm}

\def\eqref#1{equation~\ref{#1}}
\def\1{\bm{1}}

\DeclareMathAlphabet{\mathsfit}{\encodingdefault}{\sfdefault}{m}{sl}
\SetMathAlphabet{\mathsfit}{bold}{\encodingdefault}{\sfdefault}{bx}{n}

\usepackage{hyperref}
\usepackage{url}

\usepackage[utf8]{inputenc}
\usepackage[T1]{fontenc}
\usepackage{hyperref}
\usepackage{url}
\usepackage{booktabs}
\usepackage{amsfonts}
\usepackage{amsmath}
\usepackage{graphicx}
\usepackage{natbib}
\usepackage{microtype}
\usepackage{placeins}
\usepackage{multirow} % 合并行
\usepackage{adjustbox}
\usepackage{comment}
\usepackage{xcolor}

\title{Claw AI Lab: An Autonomous Multi-Agent Research Team}

\author{
\normalfont
\begin{tabular}{@{}l@{}}
{\small\bfseries
Fan Wu\textsuperscript{1,*}\quad
Cheng Chen\textsuperscript{1,2,*}\quad
Zhenshan Tan\textsuperscript{4,*}\quad
Taiyu Zhang\textsuperscript{3}\quad
Xinzhen Xu\textsuperscript{1}\quad
Yanyu Qian\textsuperscript{1}
}\\[0.3em]
{\small\bfseries
Dingcheng Gao\textsuperscript{5}\quad
Lanyun Zhu\textsuperscript{1}\quad
Qi Zhu\textsuperscript{3}\quad
Yi Tan\textsuperscript{6}\quad
Deyi Ji\textsuperscript{3}\quad
Guosheng Lin\textsuperscript{1}
}\\[0.3em]
{\small\bfseries
Tianrun Chen\textsuperscript{3,\ensuremath{\dagger}}\quad
Deheng Ye\textsuperscript{\ensuremath{\dagger}}\quad
Fayao Liu\textsuperscript{2,\ensuremath{\dagger}}
}\\[0.8em]
\multicolumn{1}{c}{
\footnotesize
\textsuperscript{1}NTU\quad
\textsuperscript{2}A*STAR\quad
\textsuperscript{3}Moxin Technology Co., LTD\quad
\textsuperscript{4}NUIST\quad
\textsuperscript{5}THU\quad
\textsuperscript{6}USTC
}\\[0.3em]
\multicolumn{1}{c}{
\footnotesize
\textsuperscript{*}Equal contribution\quad
\textsuperscript{\ensuremath{\dagger}}Corresponding authors
}\\[0.3em]
\multicolumn{1}{c}{
\small\textcolor{magenta}{\texttt{https://github.com/Claw-AI-Lab/Claw-AI-Lab}}
}
\end{tabular}
}

\iclrfinalcopy % Uncomment for camera-ready version, but NOT for submission.
\begin{document}

\maketitle
\begin{abstract}
We present Claw AI Lab, a lab-native autonomous research platform that advances automated research from a hidden prompt-to-paper pipeline into an interactive AI laboratory. Rather than centering the system around a single agent or a fixed serial workflow, we allow users to instantiate a full research team from one prompt, with customizable roles, collaborative workflows, real-time monitoring, artifact inspection, and rollback/resume control through a unified dashboard. The platform also supports distinct research modes for exploration, multi-agent discussion, and reproduction, making autonomous research substantially more steerable and laboratory-like in practice. A key practical contribution of Claw AI Lab lies in its Claw-Code Harness, which connects local codebases, datasets, and checkpoints to runnable experiments and feeds execution artifacts back into the research loop. As a result, the harness improves not only execution integration, but also experimental completion and result integrity: experiments are easier to inspect, iterate on, and faithfully transfer into final papers, reducing common failure modes such as partial runs and malformed result reporting. In our internal evaluation on five AI research case studies, using AutoResearchClaw as the baseline, Claw AI Lab is consistently preferred by AI expert judges on idea novelty, experiment completeness, and paper presentation quality. We view Claw AI Lab as an early step toward a new paradigm: \textbf{autonomous research as usable, interactive, and reliability-aware scientific infrastructure}.
\end{abstract}

\section{Introduction}
Recent progress in large language models has made autonomous research increasingly plausible. Prior systems such as AutoResearchClaw \citep{liu2026autoresearchclaw}, autoresearch \citep{karpathy2026autoresearch}, and other end-to-end research agents have demonstrated the feasibility of largely automated research workflows, in which a topic can be pushed from idea development toward experiments, analysis, and paper writing with limited human intervention \citep{lu2024aiscientist,yamada2025aiscientistv2,schmidgall2025agentlaboratory}. At the same time, recent work has expanded this space beyond one-shot paper generation, exploring multi-agent scientific collaboration, hypothesis generation, and more interactive forms of science automation \citep{gottweis2025aicoscientist,ghareeb2025robin,li2025personalizedresearchgroup}. Claw AI Lab takes a different step forward: instead of treating autonomous research primarily as automated paper production, it reframes it as the operation of an interactive AI laboratory.

This framing is central to the design of Claw AI Lab. The system is designed as a lab-native multi-agent research platform that enables users to create a full AI research lab from a single prompt, with customizable roles, collaborative workflows, and human intervention. Its interface centers the user experience around a unified dashboard with real-time event streams, multi-project monitoring, artifact inspection, and one-click rollback. Claw AI Lab also supports three distinct research modes---Explore, Discussion, and Reproduce---which move the system beyond a hidden serial pipeline and toward a more visible, collaborative, and controllable research environment. In this sense, Claw AI Lab is closer in spirit to interactive and continual science systems than to a purely offline paper-generation pipeline \citep{schmidgall2025agentlaboratory,gottweis2025aicoscientist,li2025personalizedresearchgroup}.

\begin{figure}[t]
    \centering
    \includegraphics[width=\textwidth]{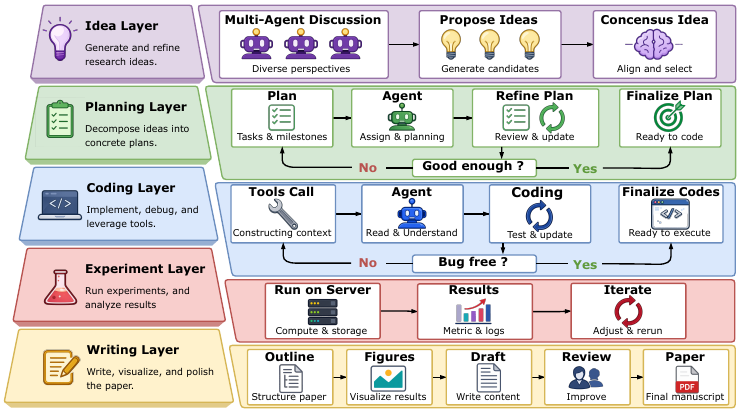}
    \caption{Overview of Claw AI Lab. The system organizes automatic research into five connected layers: idea, planning, coding, experimentation, and writing layers. Each layer uses specialized agents and validation loops, while feedback can flow across layers to revise earlier decisions when needed.}
    \label{fig:clawailab_main}
\end{figure}

This laboratory perspective is important because real research is not a one-shot generation task. It is interactive, iterative, role-specialized, and artifact-heavy. Accordingly, Claw AI Lab is designed to make autonomous research more usable in practice: users can launch projects, monitor agents, inspect intermediate artifacts, and intervene throughout the research process rather than only at the beginning or the end. In this sense, our contribution is not simply stronger automation, but a stronger systems abstraction for autonomous research---one that treats research as a persistent and inspectable process rather than a black-box pipeline.

A key practical advantage of Claw AI Lab lies in how it handles experimental execution and result consolidation. Recent systems show that coding agents can already run useful research loops over real training code and evaluation metrics \citep{karpathy2026autoresearch,zheng2025deepresearcher}. Our platform introduces Claw-Code Harness~\citep{ultraworkers2026clawcode} as a core component that reads local codebases, datasets, and checkpoints, writes runnable code, and supports the production of complete research deliverables, including papers, code, figures, and experiment logs. This design gives the harness a broader role than that of a simple execution wrapper: it becomes the interface that links local research assets to runnable experiments and connects experiment outputs back to the broader research workflow. As a result, the harness strengthens the continuity between implementation, execution, and reporting.

This point is especially important for experimental completion. In autonomous research, a common failure mode is that experiments run only partially, intermediate outputs remain difficult to inspect, or final reports contain result tables that do not faithfully reflect the actual execution outputs. Recent benchmarks suggest that multi-step research execution, replication, and evidence tracking remain significantly more difficult than surface-level generation alone might suggest \citep{starace2025paperbench,dong2026epibench}. Claw AI Lab is designed explicitly against this gap. By embedding the harness inside a dashboard-native, artifact-centered workflow, Claw AI Lab makes experimental outputs more visible, easier to trace, and easier to propagate into final reports. Put differently, the harness improves not only whether experiments can run, but whether they are carried through to complete, inspectable, and correctly reflected research artifacts. This is a central practical reason why Claw improves research completion quality in comparison to a purely pipeline-centric view of autonomous research.

Taken together, Claw AI Lab points toward a broader direction for the field. The future of autonomous research may not lie in ever longer hidden pipelines alone, but in interactive, inspectable, and reliability-aware AI laboratory systems. From this perspective, the contribution of Claw is not only a stronger platform, but a stronger framing for what autonomous research should become: not merely the automation of paper writing, but the construction of usable scientific infrastructure.

\section{Methodology}

We present Claw AI Lab, a hierarchical multi-agent framework that automates the end-to-end research process by decomposing it into five structured layers: Idea, Planning, Coding, Experiment, and Writing. 
As illustrated in the main workflow, our system mimics real-world research practices by combining role specialization, iterative refinement, and cross-stage feedback into a unified closed-loop pipeline.

\paragraph{Overview.}
Unlike prior pipeline-based research agents that operate in a linear fashion \citep{liu2026autoresearchclaw,lu2024aiscientist}, Claw AI Lab adopts a pyramid-style architecture, where high-level concepts are progressively transformed into executable artifacts. Each layer is handled by dedicated agents with distinct responsibilities, while intermediate outputs are continuously refined through validation loops. This design follows the broader move toward role-specialized research agents and interactive science automation \citep{schmidgall2025agentlaboratory,gottweis2025aicoscientist,li2025personalizedresearchgroup}. It enables both global coordination and local optimization, ensuring that early-stage decisions remain aligned with downstream execution.

\paragraph{Idea Layer.}

The process begins with a multi-agent discussion phase, where multiple agents collaboratively explore the problem space. Instead of relying on a single perspective, the system encourages diverse perspectives through parallel idea proposals, followed by structured debate and refinement. A consensus mechanism then selects and consolidates the most promising direction.
This discussion-driven design improves robustness and diversity, and better reflects how real research ideas are formed through collaboration rather than isolated generation.

\paragraph{Planning Layer.}

Given the selected idea, the system decomposes it into a structured plan consisting of tasks, dependencies, and milestones. A planning agent iteratively refines this plan through a validation loop (“Good Enough?”), where incomplete or ambiguous components are revised before execution.
Crucially, planning is not a one-shot process. Instead, it supports adaptive refinement, allowing updates based on feedback from downstream stages (e.g., coding failures or experimental results). This ensures that the plan remains feasible and aligned with practical constraints.

\paragraph{Coding Layer.}
% In the coding stage, agents translate plans into executable implementations through a tool-augmented interaction loop. Each iteration consists of context construction, code generation, tool invocation (e.g., testing, debugging), and error-driven refinement. This design is related to recent agentic research workflows that let models modify code, run experiments, and use measured results as feedback~\citep{karpathy2026autoresearch,zheng2025deepresearcher}.
% This grounding in external tools distinguishes our approach from purely generative systems, enabling reliable execution and reducing hallucinations. The process continues until the implementation satisfies correctness criteria, forming a stable bridge between abstract plans and runnable code. \todo{Claw code harness}
The Coding Layer turns an approved experiment plan into runnable research code. Centered on the Claw-Code Harness~\citep{ultraworkers2026clawcode}, it uses an agentic coding loop where the model can inspect local codebases, datasets, and checkpoints, then write, run, debug, and refine experiment files through controlled tools, including bash, read file, write file, edit file, glob search, and grep search.
The harness further improves experimental reliability by executing each task in a sandboxed workspace and injecting a read-only Python controller for time-budget enforcement, metric reporting, result finalization, and NaN/Inf detection. It also performs smoke tests and anti-fabrication checks to detect fake metrics, placeholder code, or mock implementations.
% In short, this layer converts experimental intent into executable, auditable code ready for the execution layer.

\paragraph{Experiment Layer.}
Once the implementation is finalized, the system deploys experiments on computational resources and collects metrics and logs. The experiment layer operates as an iterative optimization loop, where results are analyzed to guide subsequent adjustments.
Importantly, feedback is propagated not only within the experiment stage but also across layers. For example, unexpected results may trigger updates in the planning stage, while repeated failures may lead to revisiting the original idea. This cross-layer feedback enables continuous improvement and prevents error accumulation.

\paragraph{Writing Layer.}
The final stage transforms experimental outcomes into structured research outputs. The system generates an outline, produces visualizations, drafts the manuscript, and performs iterative review and refinement.
By integrating writing into the same pipeline, Claw AI Lab ensures consistency between experimental results and reported findings, reducing the gap between execution and documentation.

\section{Experiments}

\subsection{Experimental Setup}
\label{sec:exp_details}

Claw AI Lab is run in fully autonomous project mode. It uses GPT-5.4 as both the main model and the coding model, Gemini-3-Pro-Image-Preview as the figure-generation model for paper illustrations, and Qwen3.5-Plus/Qwen-Plus as fallback models.
AutoResearchClaw uses GPT-5.4 as the main model, Gemini-2.5-Pro-Flash-Image as the image model, and GPT-4o/GPT-4o-mini as fallback models.
We compare our method with AutoResearchClaw on four diverse topics. Topics~1--3 are research topics, while Topic~4 is a reproduction topic. The four topics are: (1) ``Quantifying Hallucination in Generated Video Models'', (2) ``LIAR Dataset-Based Fake News Classification Solution'', (3) ``A Q-Learning Approach for Student Performance Improvement Using Public Educational Data'', and (4) ``Reproducing and Analyzing PhyCustom~\citep{wu2025phycustom} on Flux~\citep{flux2024}.''
Each generated paper is reviewed by two LLM evaluators, ChatGPT 5.4 Thinking and Gemini 3.1 Pro, across six dimensions, including technical depth\&reproducibility, structure\&section flow, novelty\&contributions, clarity\&terminology, logical argumentation and citations\&evidence support.
% Each generated paper is reviewed by two LLM evaluators, ChatGPT 5.4 Thinking and Gemini 3.1 Pro in terms of \todo{}.
Each review is conducted in a fresh conversation window to reduce context carry-over. Research papers use the same academic review prompt, while the reproduction paper uses a separate reproduction-oriented prompt.

% The runtime timezone is Asia/Shanghai. The daily paper budget is 5, the quality threshold is 3.0, the time budget for each experiment is 2400 seconds, and the maximum number of iterations is 3. 
% Paper~1 uses \textit{Independent Exploration}, Papers~2--3 use \textit{Discussion Research}, and Paper~4 uses \textit{Reproduction}.

% is also run in full-auto mode. It

% The daily paper budget is 10, the quality threshold is 4.0, the time budget is 2400 seconds, and the maximum number of iterations is 10. Unlike Claw AI Lab, it mainly follows one unified multi-stage pipeline instead of task-specific research modes.

\subsection{Experimental Results}

\begin{table*}[t]
\centering
\small
\setlength{\tabcolsep}{6pt}
\begin{tabular}{l c c c c c}
\toprule
\multirow{2}*{Paper} & \multicolumn{2}{c}{ChatGPT} & \multicolumn{2}{c}{Gemini} & \multirow{2}*{Avg. Gain} \\
\cmidrule(lr){2-3}
\cmidrule(lr){4-5}
& AutoResearchClaw & Claw AI Lab & AutoResearchClaw & Claw AI Lab & \\
\midrule
Paper 1 & 62/100 & 77/100 & 68/100 & 86/100 & +16.5 \\
Paper 2 & 49/100 & 71/100 & 64/100 & 73/100 & +15.5 \\
Paper 3 & 62/100 & 73/100 & 73/100 & 95/100 & +16.5 \\
\bottomrule
\end{tabular}
\caption{Quantitative results on the three research papers of Topics 1--3 scored by different evaluators in terms of six dimensions.}
\label{tab:research_results}
\end{table*}

\begin{table}[t]
\centering
\small
\setlength{\tabcolsep}{6pt}
\begin{tabular}{l c c c}
\toprule
Evaluator & AutoResearchClaw & Claw AI Lab & Gain \\
\midrule
ChatGPT & 66/100 & 73/100 & +7.0 \\
Gemini  & 80/100 & 83/100 & +3.0 \\
\midrule
Average & 73.0/100 & 78.0/100 & +5.0 \\
\bottomrule
\end{tabular}
\caption{Quantitative result on the reproduction report of Topic 4.}
\label{tab:reproduction_results}
\end{table}

\begin{figure*}[t]
    \centering

    \begin{minipage}[t]{0.42\textwidth}
        \centering
        \includegraphics[width=\linewidth]{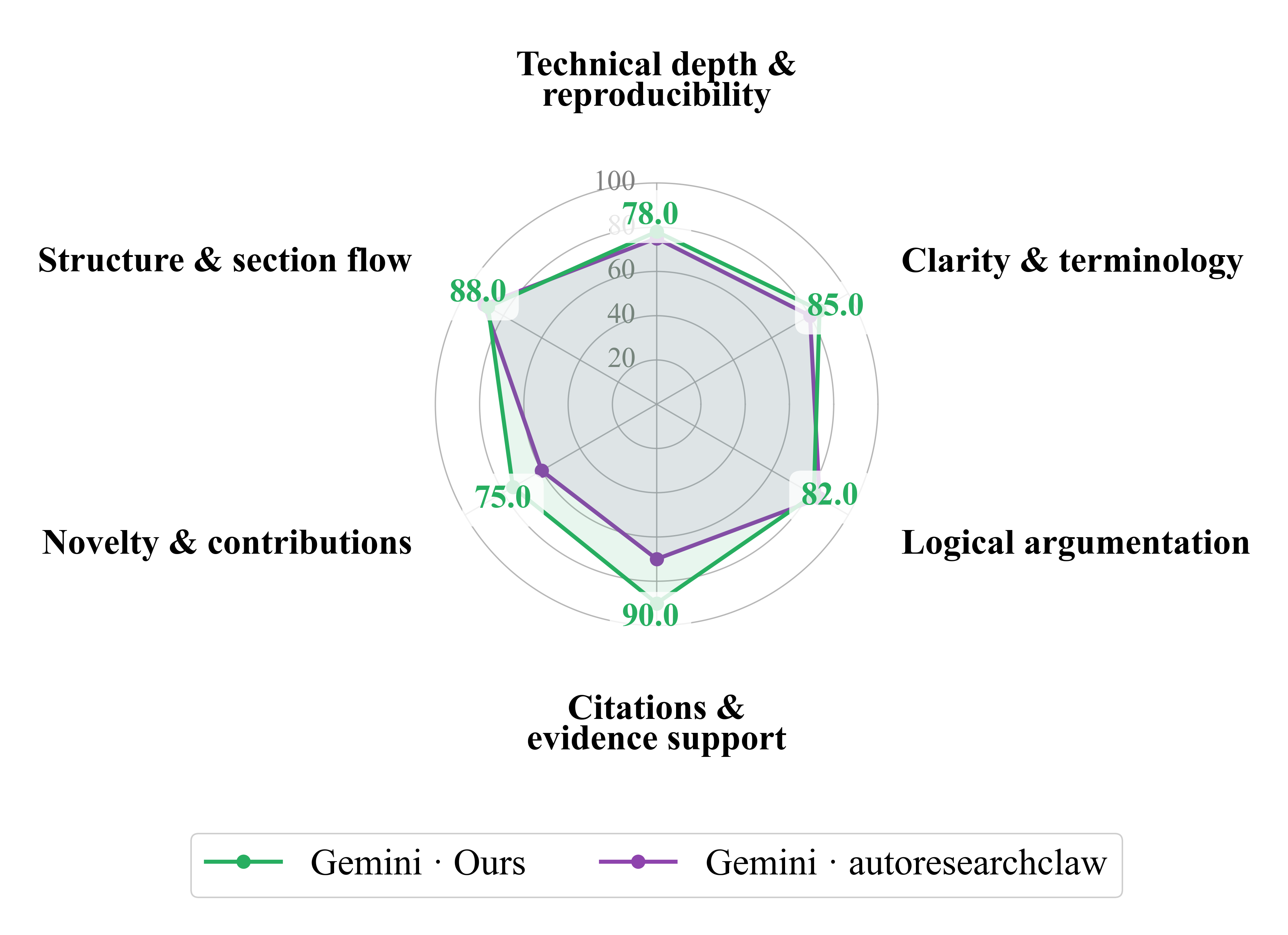}\\
        \small (a) Paper 1 scored by Gemini
    \end{minipage}
    \hfill
    \begin{minipage}[t]{0.42\textwidth}
        \centering
        \includegraphics[width=\linewidth]{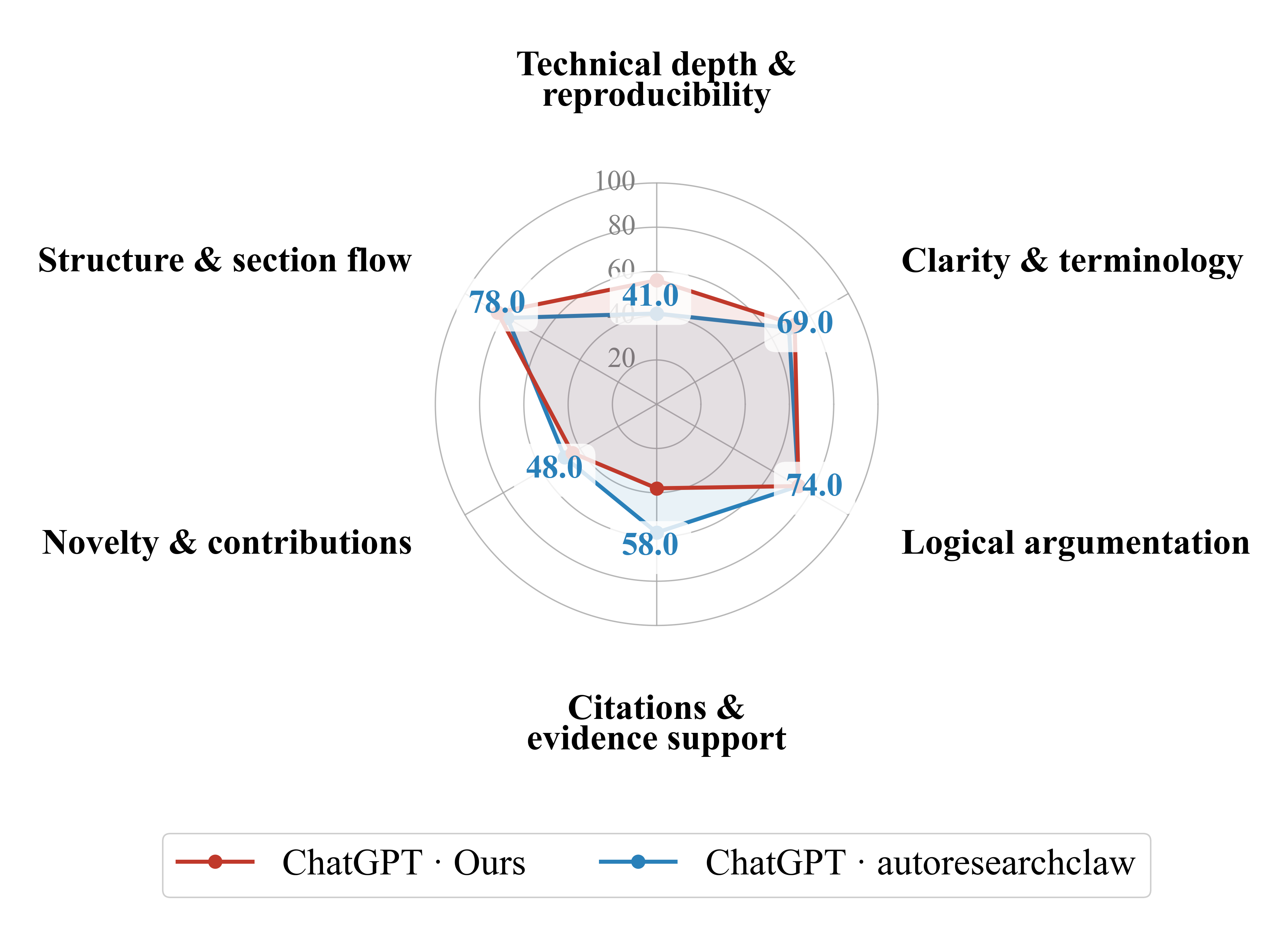}\\
        \small (b) Paper 1 scored by ChatGPT
    \end{minipage}

    \vspace{0.4em}

    \begin{minipage}[t]{0.42\textwidth}
        \centering
        \includegraphics[width=\linewidth]{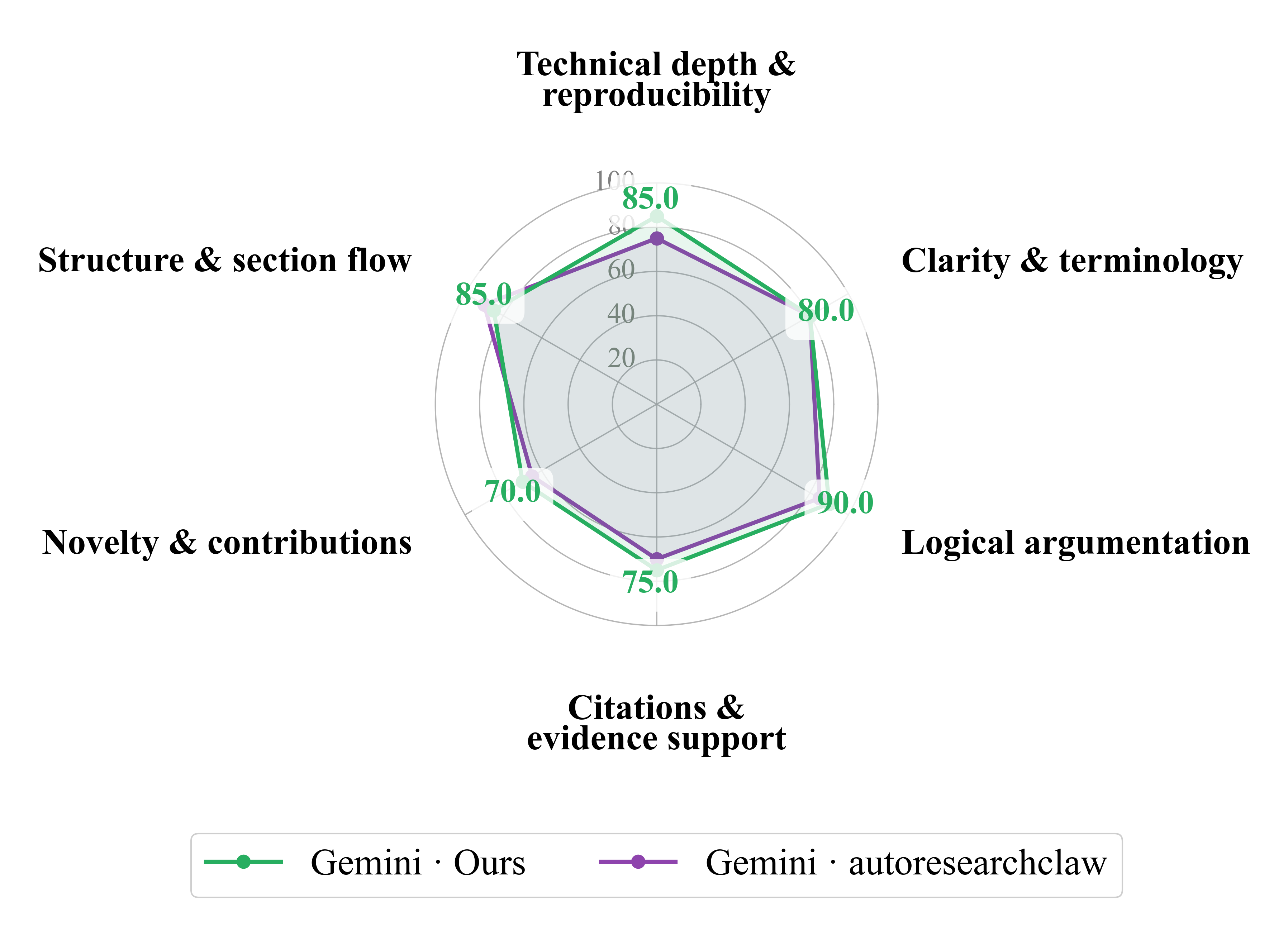}\\
        \small (c) Paper 2 scored by Gemini
    \end{minipage}
    \hfill
    \begin{minipage}[t]{0.42\textwidth}
        \centering
        \includegraphics[width=\linewidth]{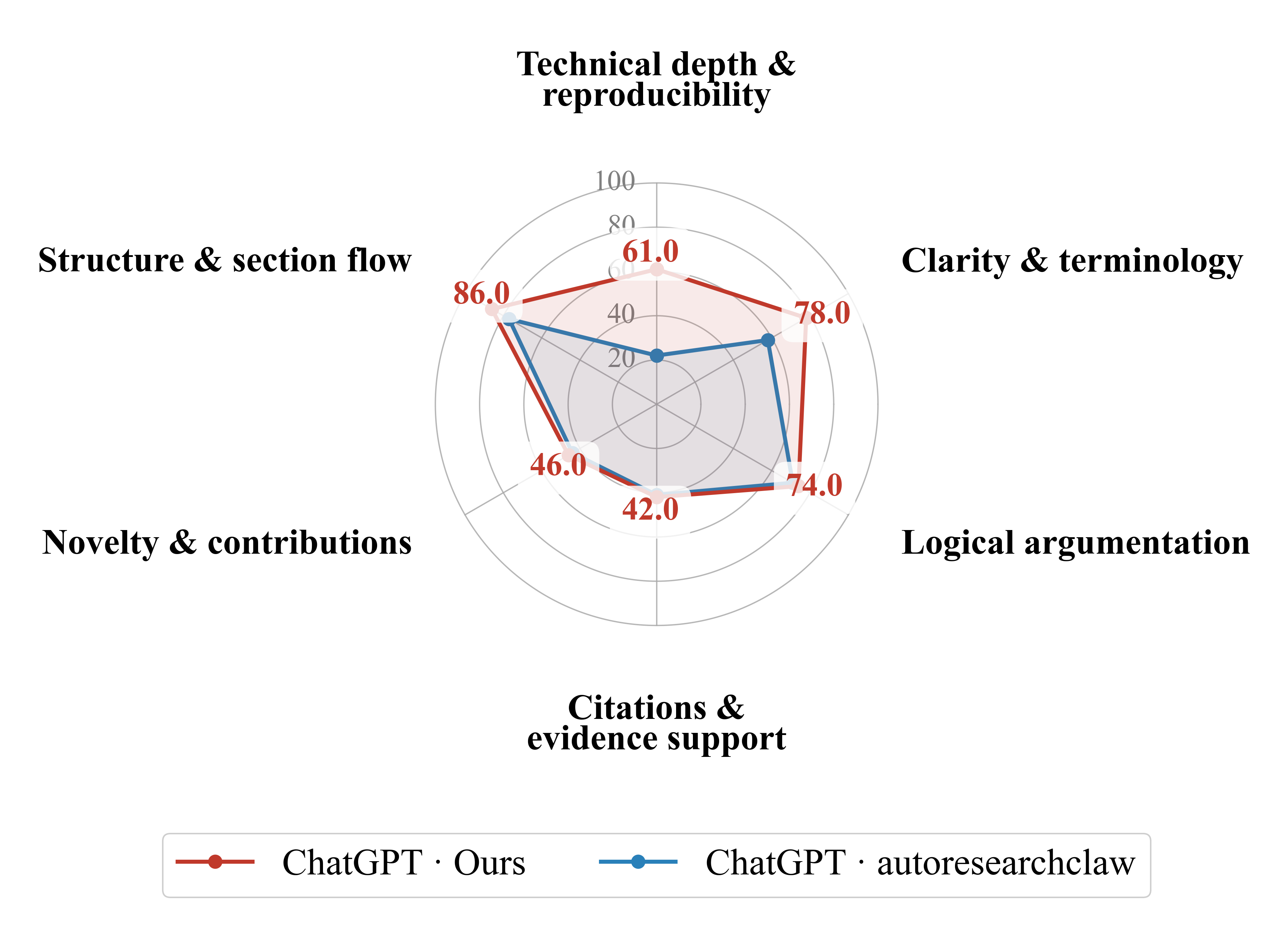}\\
        \small (d) Paper 2 scored by ChatGPT
    \end{minipage}

    \vspace{0.4em}

    \begin{minipage}[t]{0.42\textwidth}
        \centering
        \includegraphics[width=\linewidth]{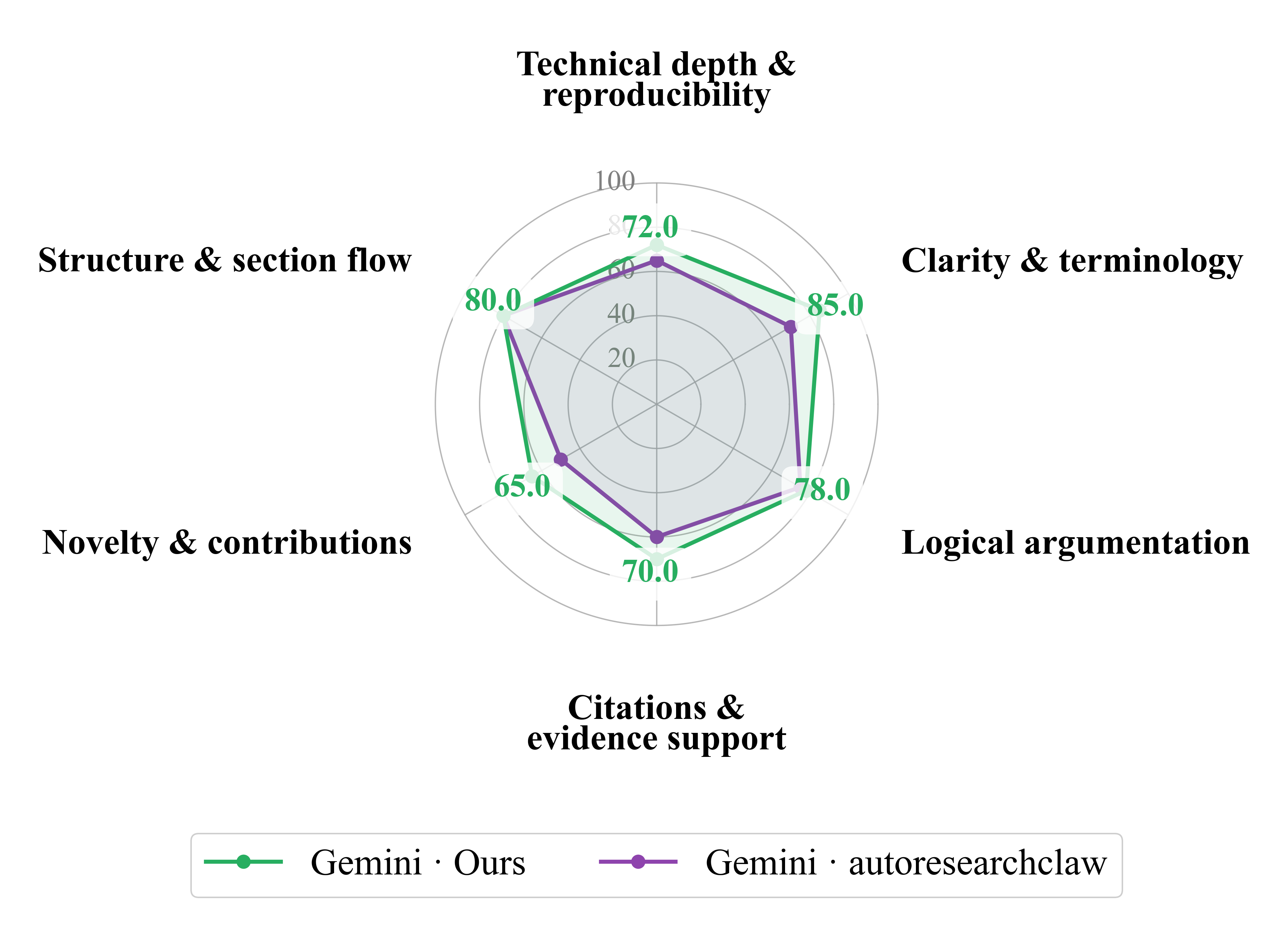}\\
        \small (e) Paper 3 scored by Gemini
    \end{minipage}
    \hfill
    \begin{minipage}[t]{0.42\textwidth}
        \centering
        \includegraphics[width=\linewidth]{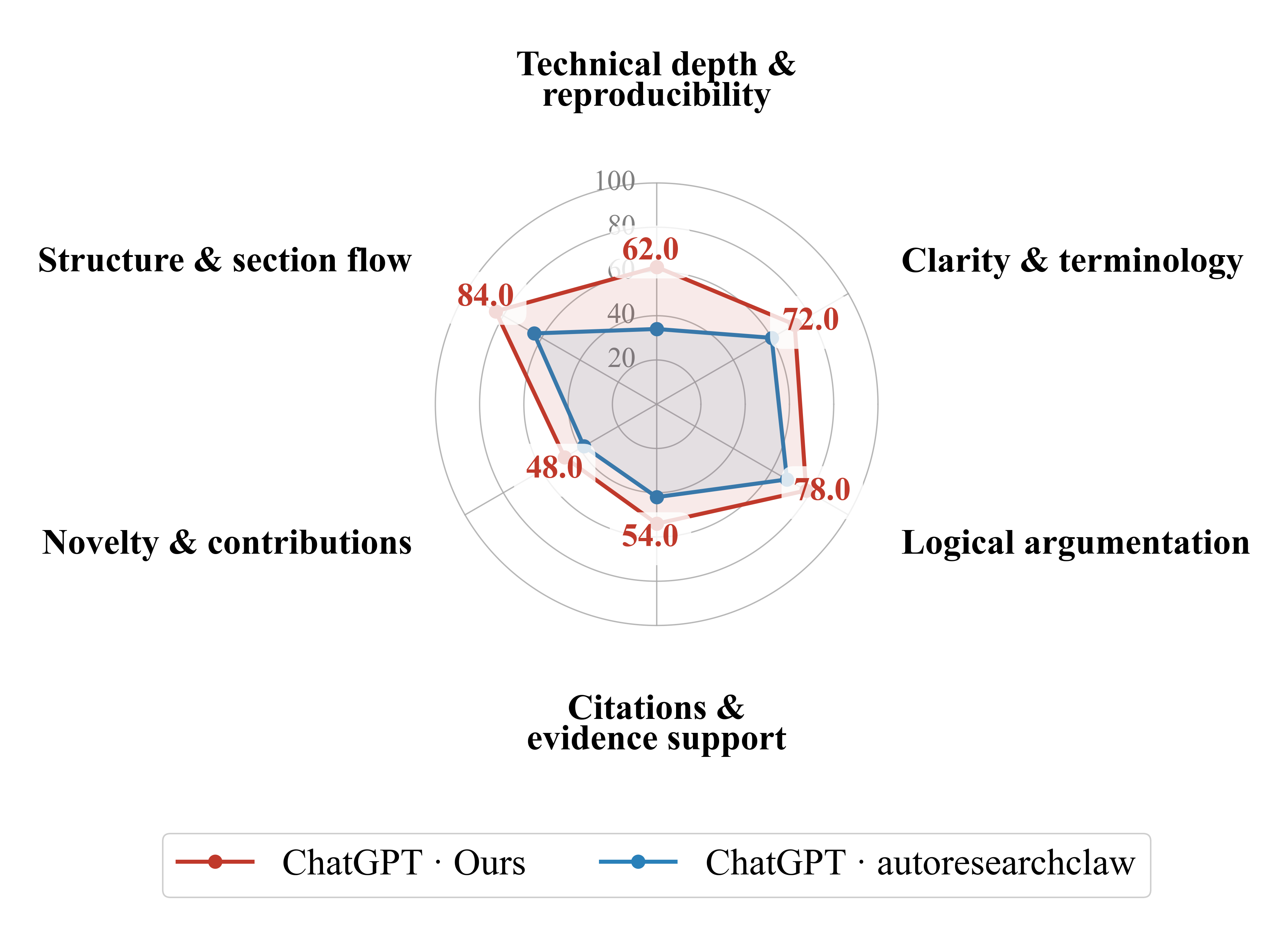}\\
        \small (f) Paper 3 scored by ChatGPT
    \end{minipage}

    \vspace{0.4em}

    \begin{minipage}[t]{0.42\textwidth}
        \centering
        \includegraphics[width=\linewidth]{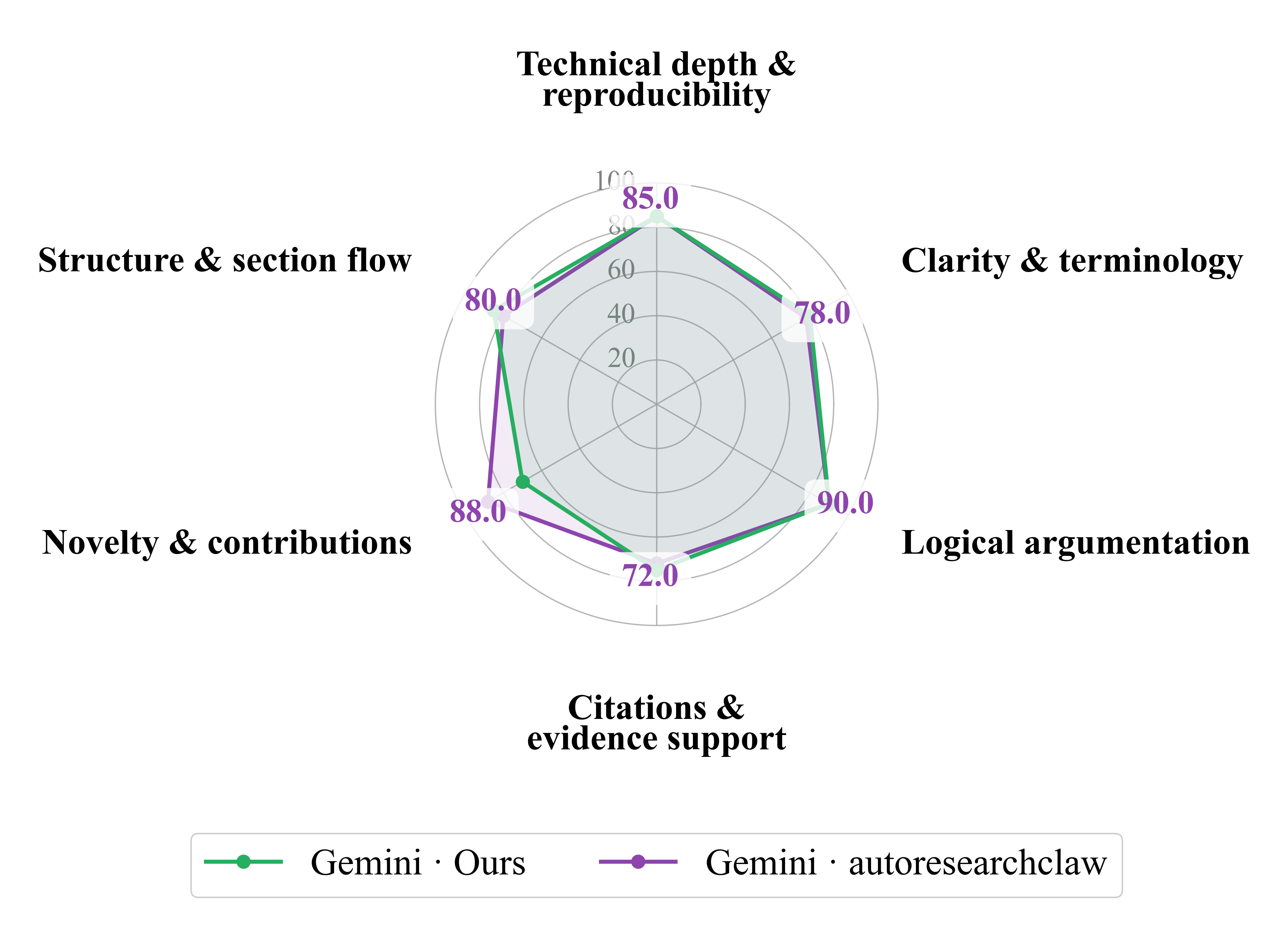}\\
        \small (g) Paper 4 scored by Gemini
    \end{minipage}
    \hfill
    \begin{minipage}[t]{0.42\textwidth}
        \centering
        \includegraphics[width=\linewidth]{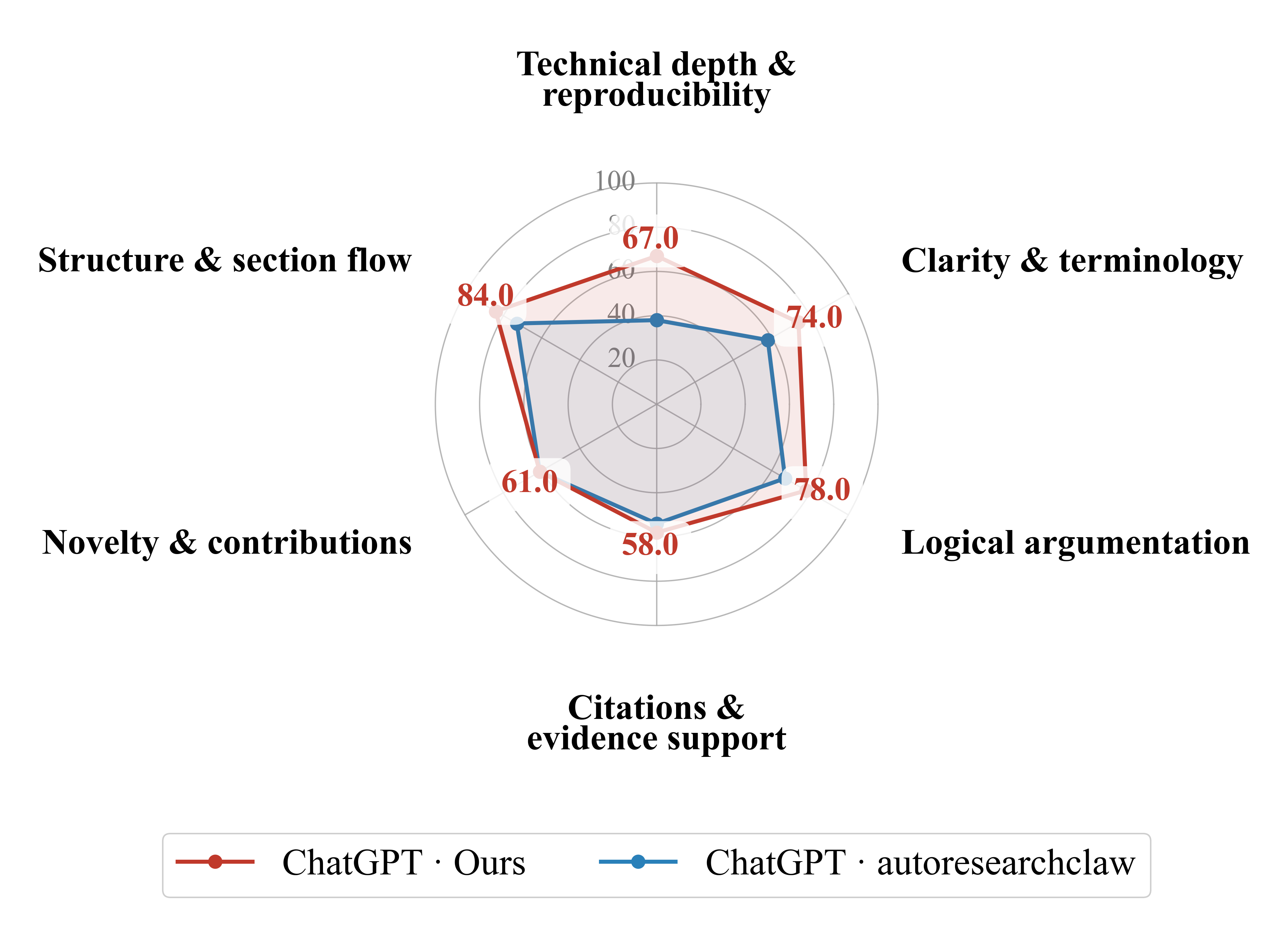}\\
        \small (h) Paper 4 scored by ChatGPT
    \end{minipage}

    \caption{Detailed comparison for four paper pairs scored by Gemini and ChatGPT, respectively. Each panel compares Claw AI Lab with AutoResearchClaw on six dimensions.}
    \label{fig:radar_four_cases}
\end{figure*}

Tables~\ref{tab:research_results} and~\ref{tab:reproduction_results} summarize the quantitative comparison between Claw AI Lab and AutoResearchClaw across four topics and two evaluators. 
As shown in Table~\ref{tab:research_results}, Claw AI Lab achieves consistent gains across Topics~1--3, with average improvements ranging from +15.5 to +16.5 points. 
For the reproduction topic, Table~\ref{tab:reproduction_results} shows that the average score increases from 73.0/100 to 78.0/100, corresponding to a 5.0-point improvement. 
Both ChatGPT and Gemini evaluators consistently assign higher scores to Claw AI Lab than to AutoResearchClaw across all topics, indicating that the improvement is stable across different evaluation protocols. 
Furthermore, Figure~\ref{fig:radar_four_cases} provides a fine-grained comparison on six dimensions. Overall, Claw AI Lab exhibits a more competitive and balanced performance across most cases. 
These results suggest that Claw AI Lab benefits from our more reliable and efficient Claw-Code harness, which enables more trustworthy experimental execution and provides stronger empirical support for the generated papers, thereby leading to more stable improvements in overall paper quality.

\clearpage

\bibliography{iclr2026_conference}

\begin{thebibliography}{14}
\providecommand{\natexlab}[1]{#1}
\providecommand{\url}[1]{\texttt{#1}}
\expandafter\ifx\csname urlstyle\endcsname\relax
  \providecommand{\doi}[1]{doi: #1}\else
  \providecommand{\doi}{doi: \begingroup \urlstyle{rm}\Url}\fi

\bibitem[Dong et~al.(2026)Dong, Zheng, Niu, Han, Li, Liu, Liu, Li, Zhu, and Che]{dong2026epibench}
Xuan Dong, Huanyang Zheng, Tianhao Niu, Zhe Han, Pengzhan Li, Bofei Liu, Zhengyang Liu, Guancheng Li, Qingfu Zhu, and Wanxiang Che.
\newblock Epibench: Benchmarking multi-turn research workflows for multimodal agents.
\newblock \emph{arXiv preprint arXiv:2604.05557}, 2026.

\bibitem[Ghareeb et~al.(2025)Ghareeb, Chang, Mitchener, Yiu, Szostkiewicz, Laurent, Razzak, White, Hinks, and Rodriques]{ghareeb2025robin}
Ali~Essam Ghareeb, Benjamin Chang, Ludovico Mitchener, Angela Yiu, Caralyn~J. Szostkiewicz, Jon~M. Laurent, Muhammed~T. Razzak, Andrew~D. White, Michaela~M. Hinks, and Samuel~G. Rodriques.
\newblock Robin: A multi-agent system for automating scientific discovery.
\newblock \emph{arXiv preprint arXiv:2505.13400}, 2025.

\bibitem[Gottweis et~al.(2025)Gottweis, Weng, Daryin, Tu, Palepu, Sirkovic, Myaskovsky, Weissenberger, Rong, Tanno, et~al.]{gottweis2025aicoscientist}
Juraj Gottweis, Wei-Hung Weng, Alexander Daryin, Tao Tu, Anil Palepu, Petar Sirkovic, Artiom Myaskovsky, Felix Weissenberger, Keran Rong, Ryutaro Tanno, et~al.
\newblock Towards an ai co-scientist.
\newblock \emph{arXiv preprint arXiv:2502.18864}, 2025.

\bibitem[Karpathy(2026)]{karpathy2026autoresearch}
Andrej Karpathy.
\newblock autoresearch, 2026.
\newblock URL \url{https://github.com/karpathy/autoresearch}.

\bibitem[Labs(2024)]{flux2024}
Black~Forest Labs.
\newblock Flux.
\newblock \url{https://github.com/black-forest-labs/flux}, 2024.

\bibitem[Li et~al.(2025)Li, Ren, Pan, Yan, Li, Bergemann, and Yang]{li2025personalizedresearchgroup}
Ed~Li, Junyu Ren, Xintian Pan, Cat Yan, Chuanhao Li, Dirk Bergemann, and Zhuoran Yang.
\newblock Build your personalized research group: A multiagent framework for continual and interactive science automation.
\newblock \emph{arXiv preprint arXiv:2510.15624}, 2025.

\bibitem[Liu et~al.(2026)Liu, Xia, Han, Qiu, Zhang, Chen, Tu, Yang, Zhou, Zhu, Li, Zhang, Zhou, Zheng, Xie, Ding, and Yao]{liu2026autoresearchclaw}
Jiaqi Liu, Peng Xia, Siwei Han, Shi Qiu, Letian Zhang, Guiming Chen, Haoqin Tu, Xinyu Yang, Jiawei Zhou, Hongtu Zhu, Yun Li, Jiaheng Zhang, Yuyin Zhou, Zeyu Zheng, Cihang Xie, Mingyu Ding, and Huaxiu Yao.
\newblock Autoresearchclaw: Fully autonomous research from idea to paper, 2026.
\newblock URL \url{https://github.com/aiming-lab/AutoResearchClaw}.

\bibitem[Lu et~al.(2024)Lu, Lu, Lange, Foerster, Clune, and Ha]{lu2024aiscientist}
Chris Lu, Cong Lu, Robert~Tjarko Lange, Jakob~N. Foerster, Jeff Clune, and David Ha.
\newblock The ai scientist: Towards fully automated open-ended scientific discovery.
\newblock \emph{arXiv preprint arXiv:2408.06292}, 2024.

\bibitem[Schmidgall et~al.(2025)Schmidgall, Su, Wang, Sun, Wu, Yu, Liu, Moor, Liu, and Barsoum]{schmidgall2025agentlaboratory}
Samuel Schmidgall, Yusheng Su, Ze~Wang, Ximeng Sun, Jialian Wu, Xiaodong Yu, Jiang Liu, Michael Moor, Zicheng Liu, and Emad Barsoum.
\newblock Agent laboratory: Using llm agents as research assistants.
\newblock In \emph{Findings of the Association for Computational Linguistics: EMNLP 2025}, pp.\  5977--6043, 2025.

\bibitem[Starace et~al.(2025)Starace, Jaffe, Sherburn, Aung, Chan, Maksin, Dias, Mays, Kinsella, Thompson, Heidecke, Glaese, and Patwardhan]{starace2025paperbench}
Giulio Starace, Oliver Jaffe, Dane Sherburn, James Aung, Jun~Shern Chan, Leon Maksin, Rachel Dias, Evan Mays, Benjamin Kinsella, Wyatt Thompson, Johannes Heidecke, Amelia Glaese, and Tejal Patwardhan.
\newblock Paperbench: Evaluating ai's ability to replicate ai research.
\newblock In \emph{Proceedings of the 42nd International Conference on Machine Learning}, volume 267 of \emph{Proceedings of Machine Learning Research}, pp.\  56843--56873, 2025.

\bibitem[{UltraWorkers}(2026)]{ultraworkers2026clawcode}
{UltraWorkers}.
\newblock {Claw Code}.
\newblock \url{https://github.com/ultraworkers/claw-code}, 2026.
\newblock Public Rust implementation of the \texttt{claw} CLI agent harness. Accessed: 2026-05-18.

\bibitem[Wu et~al.(2025)Wu, Chen, Fu, Wei, Xu, Ye, and Lin]{wu2025phycustom}
Fan Wu, Cheng Chen, Zhoujie Fu, Jiacheng Wei, Yi~Xu, Deheng Ye, and Guosheng Lin.
\newblock Phycustom: Towards realistic physical customization in text-to-image generation.
\newblock \emph{arXiv preprint arXiv:2512.02794}, 2025.

\bibitem[Yamada et~al.(2025)Yamada, Lange, Lu, Hu, Lu, Foerster, Clune, and Ha]{yamada2025aiscientistv2}
Yutaro Yamada, Robert~Tjarko Lange, Cong Lu, Shengran Hu, Chris Lu, Jakob Foerster, Jeff Clune, and David Ha.
\newblock The ai scientist-v2: Workshop-level automated scientific discovery via agentic tree search.
\newblock \emph{arXiv preprint arXiv:2504.08066}, 2025.

\bibitem[Zheng et~al.(2025)Zheng, Fu, Hu, Cai, Ye, Lu, and Liu]{zheng2025deepresearcher}
Yuxiang Zheng, Dayuan Fu, Xiangkun Hu, Xiaojie Cai, Lyumanshan Ye, Pengrui Lu, and Pengfei Liu.
\newblock Deepresearcher: Scaling deep research via reinforcement learning in real-world environments.
\newblock In \emph{Proceedings of the 2025 Conference on Empirical Methods in Natural Language Processing}, pp.\  414--431, 2025.

\end{thebibliography}
\bibliographystyle{iclr2026_conference}

% \appendix
% \section{Appendix}

\end{document}